\documentclass[conference]{IEEEtran}
\IEEEoverridecommandlockouts
\usepackage{cite}
\usepackage{amsmath,amssymb,amsfonts}
\usepackage{algorithm}
\usepackage{algpseudocode}
\usepackage{graphicx}
\usepackage{textcomp}
\usepackage{xcolor}
\usepackage{subcaption}
\usepackage{booktabs}
\graphicspath{{ }}
\def\BibTeX{{\rm B\kern-.05em{\sc i\kern-.025em b}\kern-.08em
    T\kern-.1667em\lower.7ex\hbox{E}\kern-.125emX}}
\makeatletter
\newcommand{\linebreakand}{%
  \end{@IEEEauthorhalign}
  \hfill\mbox{}\\
  \mbox{}\hfill\begin{@IEEEauthorhalign}
}
\makeatother
\begin{document}

\title{Agent2UCB: Agentic System for Generative Engine Optimization}

\author{\IEEEauthorblockN{Sheldon Yu}
\IEEEauthorblockA{\textit{UC San Diego} \\
La Jolla, CA, USA \\
ziy040@ucsd.edu}
\and
\IEEEauthorblockN{Rui Wang}
\IEEEauthorblockA{\textit{Adobe Research} \\
San Jose, CA, USA \\
ruiwan@adobe.com}
\and
\IEEEauthorblockN{Tong Yu}
\IEEEauthorblockA{\textit{Adobe Research} \\
San Jose, CA, USA \\
tyu@adobe.com}
\linebreakand
\IEEEauthorblockN{Sungchul Kim}
\IEEEauthorblockA{\textit{Adobe Research} \\
San Jose, CA, USA \\
sukim@adobe.com}
\and
\IEEEauthorblockN{Doga Dogan}
\IEEEauthorblockA{\textit{Adobe Research} \\
San Jose, CA, USA \\
doga@adobe.com}
\and
\IEEEauthorblockN{Junda Wu}
\IEEEauthorblockA{\textit{Adobe Research} \\
San Jose, CA, USA \\
jundaw@adobe.com}
\and
\IEEEauthorblockN{Julian McAuley}
\IEEEauthorblockA{\textit{UC San Diego} \\
La Jolla, CA, USA \\
jmcauley@ucsd.edu}
}

\maketitle
\begin{abstract}
Large language model driven search engines such as Google AI Overviews and Perplexity have created new opportunities for \textbf{Generative Engine Optimization (GEO)}---the practice of refining content to increase its likelihood of being cited or summarized by generative systems. We demonstrate \textbf{Agent2UCB}, an agentic GEO system that autonomously improves content visibility through customized, feedback-driven optimization. For each content item, the system evaluates nine GEO strategies, identifies the most effective method, and accelerates selection using a \textbf{bandit-based Agent2UCB} policy that integrates LLM priors with online reward signals. To monitor side effects, the system also provides a lightweight, text-only \textbf{SEO readiness evaluation} covering readability, topical coverage, and EEAT-style credibility. Experiments on GEO-Bench show consistent visibility gains while preserving SEO quality. The demo allows users to choose the websites of interest, observe the optimization workflow and compare GEO/SEO outcomes across methods.
\end{abstract}

\section{Introduction}
\begin{figure*}[t]
    \centering
    \rotatebox{270}{
\includegraphics[width=\textwidth,height=0.80\textheight,keepaspectratio]{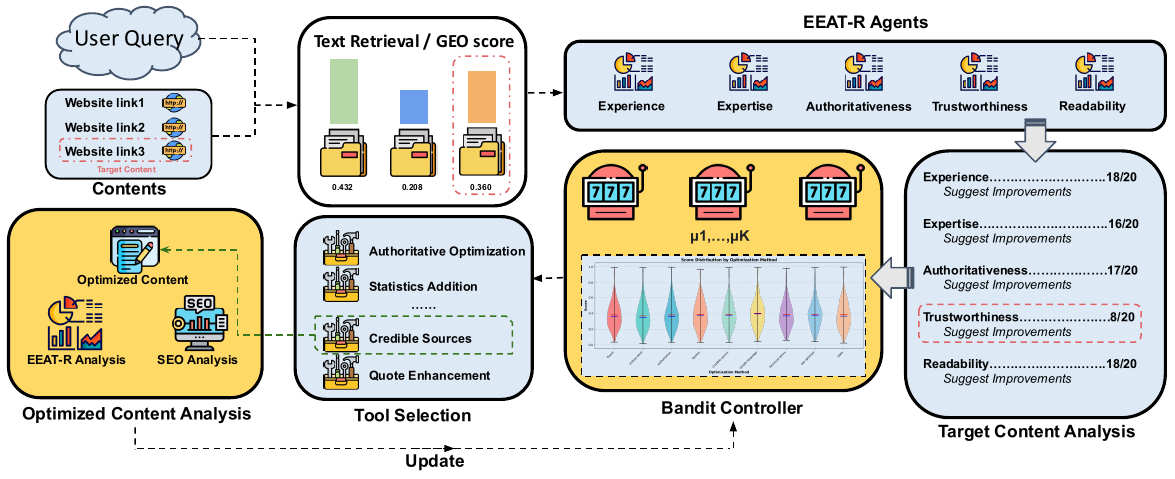}
    }
    \caption{Overview of the Agent2UCB pipeline. The Text Retrieval / GEO score block computes baseline generative-visibility scores for each candidate URL; the Target content analysis block runs EEAT-R diagnostics on the selected page; the Tool selection block lets the bandit controller apply a GEO optimization tool to produce a new draft; and the Optimized content analysis block re-scores that draft on both GEO and SEO readiness, closing the loop back to the user-facing dashboard.}
    \label{fig:overview}
\end{figure*}

While LLM-powered features such as Google AI Overviews, Perplexity, and ChatGPT’s deep research are increasingly surfacing web content in generative answers, site operators currently have little control over how their pages are selected and cited. Early work on Generative Engine Optimization has shown that rewriting a page can change its likelihood of being used by generative systems, but the tooling remains primitive compared to traditional SEO: most pipelines apply a single, category-level “best” rewrite strategy, ignore item level variation, and offer no feedback beyond aggregate benchmark scores. In practice, content owners need a system that can adaptively choose optimization methods for each page, expose the trade offs between generative visibility and SEO style quality, and do so under realistic budget constraints on LLM calls.

However, existing GEO pipelines\cite{GEO} are largely \emph{generic and non-generalized}. Benchmarks such as the KDD 2024 GEO suite typically identify a single, \emph{category-level} optimization strategy that performs best on average within a topical group~\cite{GEO}. While such methods establish valuable baselines, they overlook the fact that the most effective optimization method can vary dramatically across individual content items even within the same category. In practice, current systems either (i) apply a fixed, one-size-fits-all strategy per category, or (ii) rely on exhaustive evaluation, testing all available optimization tools for every query and then picking the best. Both patterns limit generalization and are computationally expensive for large-scale or interactive deployment.

At the same time, GEO optimization today is static and offline. Once a strategy is chosen, it is applied uniformly without leveraging \textit{verifiable feedback} from generative engines or offline simulators. Recent advances in \textbf{agentic and feedback driven optimization}~\cite{arumugam2025exploration} and \textbf{bandit and adaptive control for LLM systems}~\cite{bouneffouf2025bandits,nie2024evolve} suggest a more powerful alternative: agentic controllers that iteratively try different tools, observe visibility gains, and adapt their decisions over time using multi-armed bandit policies~\cite{auer2002finite}. Yet these ideas have mostly appeared as algorithms and simulation studies, rather than as deployable systems that practitioners can experiment with and integrate into their workflows.

A further gap lies in the \emph{human-facing analytics} for GEO. Traditional SEO already offers mature tooling, such as Lighthouse~\cite{lighthouse}, Search Console~\cite{searchconsole}, and third party dashboards—that explain why a page performs well or poorly and suggest actionable improvements. In contrast, existing GEO work typically exposes only offline evaluation scripts and aggregate metrics. There is, to our knowledge, no end-to-end GEO system that (i) runs generative optimization, (ii) selects among multiple optimization tools dynamically, and (iii) surfaces \emph{interpretable analyses} and recommendations to human users in a Lighthouse-style interface, including both GEO and SEO signals.

In this demo paper, we present \textbf{Agent2UCB}, an experimentation-augmented GEO system that turns these ideas into a practical, interactive tool. Agent2UCB treats each content item as a self-contained optimization task: it applies nine GEO optimization strategies, obtains verifiable feedback from a GEO-style evaluator, and learns to select the most effective method for that specific item. To avoid brute-force exploration, the system employs a \textbf{bandit-based Agent2UCB controller} that integrates LLM-based priors with online visibility feedback, achieving high GEO performance within only a small number of iterations. On top of this optimization engine, Agent2UCB provides a \textbf{GEO+SEO analytics dashboard} that reports generative visibility, a lightweight text-only SEO readiness score, and tool-level attribution, giving practitioners a GEO analogue to SEO Lighthouse. We summarize our contributions as follows:
\begin{itemize}
    \item \textbf{Customized Agentic Optimization.} We implement an agentic GEO engine that automatically applies and evaluates nine optimization tools per content item, selecting the best-performing method through feedback-driven comparison. This enables per-item customization beyond prior category-level GEO baselines.
    \item \textbf{Bandit-Enhanced Efficiency.} We integrate a \textit{bandit-based Agent2UCB} controller that models each optimization method as an arm and uses LLM-informed priors with online visibility rewards to balance exploration and exploitation, identifying effective tools within about 15 iterations, $10\times$ more efficient than brute-force search.
    \item \textbf{Human-Centric GEO + SEO Analytics.} We build a text-only SEO readiness evaluation pipeline that quantifies readability, topical coverage, and credibility, and expose these signals alongside GEO scores in an interactive dashboard. This allows users to inspect how optimization affects both generative visibility and SEO-style quality, closing the loop between automated GEO and human decision-making.
\end{itemize}
Together, these contributions introduce a scalable and adaptive framework for real-time content optimization in generative search, and demonstrate a practical, Lighthouse-style system that bridges agentic reasoning, online learning, and human-centric analytics for GEO.

\section{System Architecture}

\begin{figure}[t]
    \centering

    % -------- Top Row (two images) --------
    \begin{subfigure}{0.48\columnwidth}
        \centering
        \includegraphics[width=\linewidth]{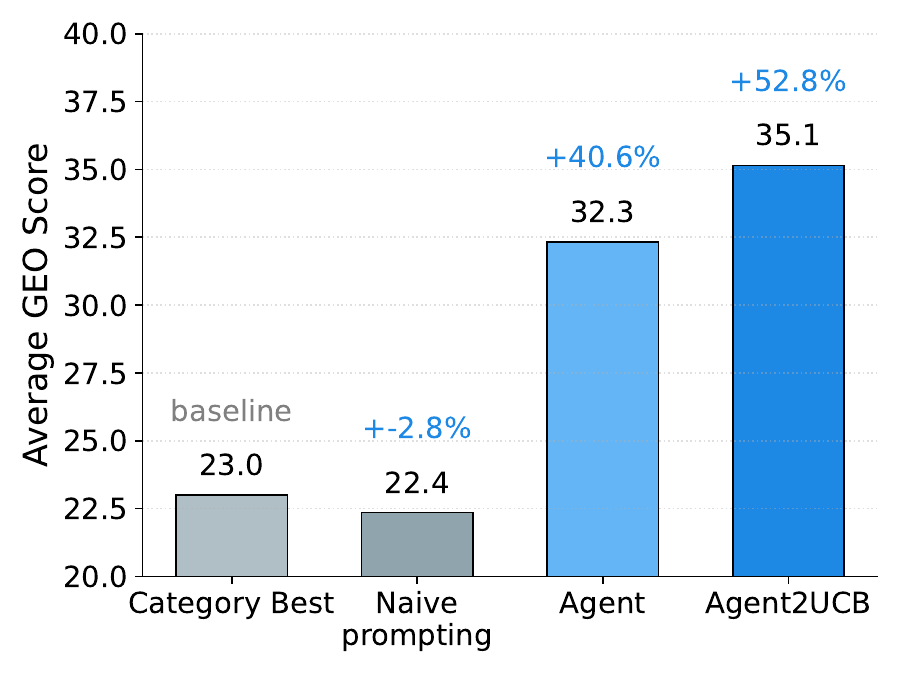}
    \end{subfigure}
    \hfill
    \begin{subfigure}{0.48\columnwidth}
        \centering
        \includegraphics[width=\linewidth]{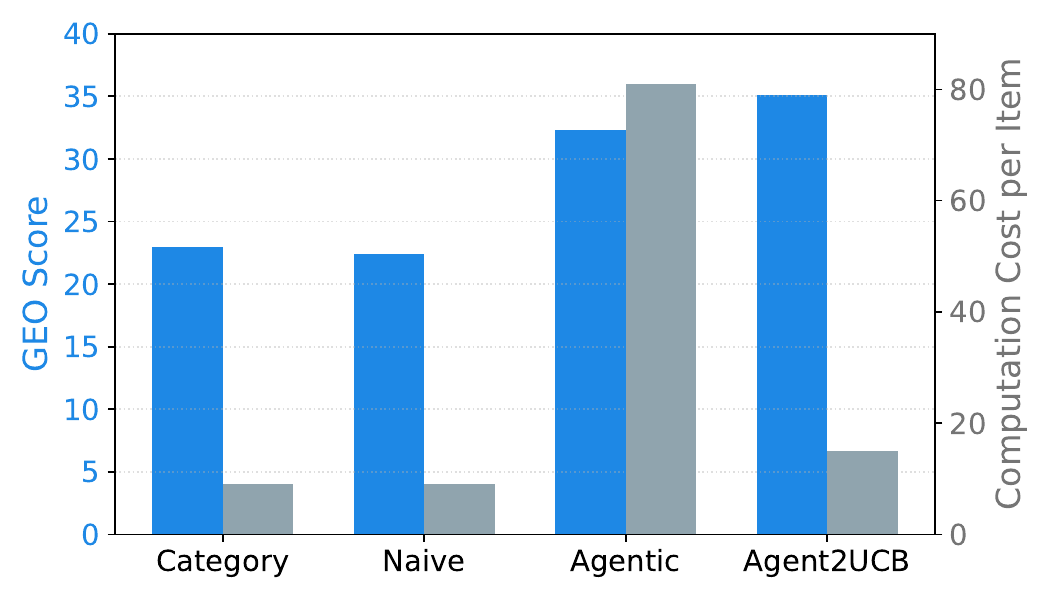}
    \end{subfigure}

    % -------- Bottom Row (one image full-width) --------
    \vspace{0.5em} % spacing between rows
    \begin{subfigure}{0.98\columnwidth}
        \centering
        \includegraphics[width=\linewidth]{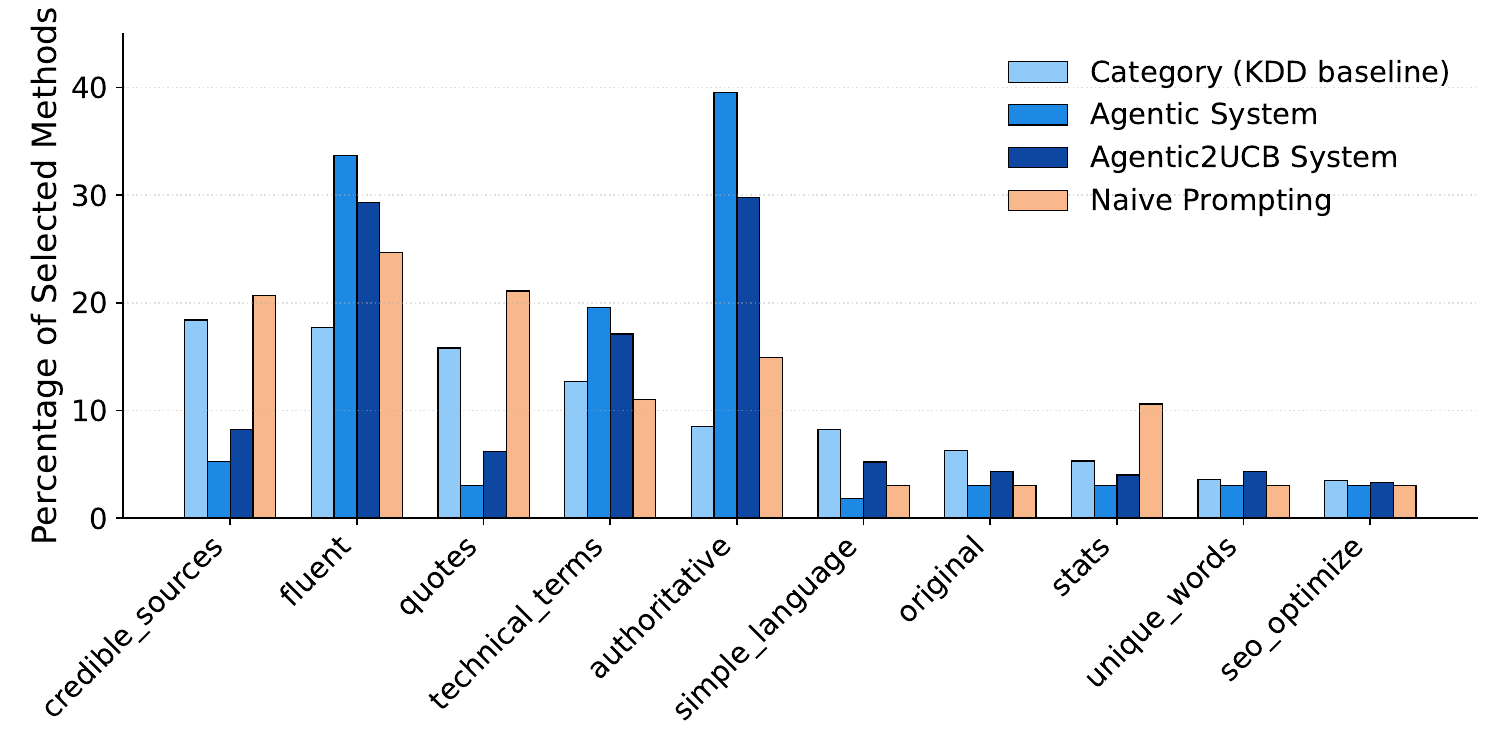}
    \end{subfigure}

    \caption{RQ1: average GEO score by method (left); RQ2: GEO score vs.\ bandit iterations (right); distribution of tools selected across content items (bottom).}
    \label{fig:bandit_results}
\end{figure}

Agent2UCB takes as input a user query and a set of website links and outputs optimized content together with GEO and SEO-style analytics. As illustrated in Figure~\ref{fig:overview}, the system first analyzes how each page behaves under generative search, then diagnoses its EEAT-R profile, and finally applies bandit-guided optimization tools to the selected target page. The loop repeats until a high-performing version is found, which is then surfaced to the user through a Lighthouse-style dashboard. The system is organized into four main components:

\begin{enumerate}
    \item \textbf{Input and GEO scoring.}
    The system takes a single user query and multiple website links that the user wishes to compare or optimize. Given a query $q$ and a set of URLs $\{u_1, \dots, u_M\}$, Agent2UCB retrieves each page $u_m$, extracts a clean text representation with an LLM-based extractor, and runs a GEO evaluator that simulates generative search: an LLM is prompted to answer the query using all candidate pages, and we measure how often and how prominently each page is cited. These signals are aggregated into a GEO score $\mathrm{GEO}(q, d_m)$ for each URL (Text Retrieval / GEO score block in Figure~\ref{fig:overview}), providing the baseline visibility against which future improvements are measured.

    \item \textbf{EEAT-R analysis.}
    For the chosen target page, a text-only SEO readiness analyzer scores the content along five dimensions: Experience, Expertise, Authoritativeness, Trustworthiness, and Readability (EEAT-R)~\cite{googleeeat}. The resulting scores and explanations (Target content analysis in Figure~\ref{fig:overview}) highlight strengths and weaknesses, such as low trustworthiness, and generate concrete natural-language suggestions (e.g., add citations, clarify authorship, simplify language). These insights guide which optimization tools are likely to be effective.

    \item \textbf{Agentic GEO engine and tool library.}
    At the core of Agent2UCB is an agentic GEO engine that treats each content item as its own optimization task. Given the query $q$ and target document $d$, the engine coordinates a library of $K$ GEO tools $\{a_1, \dots, a_K\}$, including:
    \begin{itemize}
        \item authoritative rewriting to strengthen expertise,
        \item evidence injection to increase factual grounding,
        \item quote and citation to highlight credible sources,
        \item readability tuning for clearer explanations,
        \item SEO-friendly structuring.
    \end{itemize}
    Each tool $a_k$ is an LLM prompt template that takes $(q, d)$ and produces an optimized draft $d^{(k)}$ (Tool selection block in Figure~\ref{fig:overview}). When a tool is applied, the GEO evaluator re-runs the generative search simulation on $d^{(k)}$ and the SEO readiness module recomputes its metrics, exposing any gains or regressions to the user. In a purely exhaustive pipeline, the system would apply all $K$ tools to every item before choosing the best; Agent2UCB instead lets its bandit controller decide which tools to try and when to stop exploring, reducing the number of LLM calls per item.

    \item \textbf{Agent2UCB Bandit Controller.}
    To make tool selection adaptive and sample-efficient, Agent2UCB models GEO optimization as a multi-armed bandit problem. For a fixed query--document pair $(q, d)$, each tool $a_k$ is an arm with an (unknown) expected reward $\mathbb{E}[r_k]$, the observed visibility gain. We adopt an upper confidence bound (UCB) style policy~\cite{auer2002finite}, where the score for arm $k$ at iteration $t$ is
    \[
        \mathrm{UCB}_k(t) = \hat{\mu}_k(t) + \alpha \sqrt{\frac{\log t}{n_k(t)}},
    \]
    with $\hat{\mu}_k(t)$ the empirical mean reward, $n_k(t)$ the selection count, and $\alpha > 0$ a tunable exploration parameter. Beyond standard UCB, Agent2UCB incorporates \emph{LLM-based priors} by querying a lightweight tool-selection agent before online interaction; in the dual-UCB variant used in our system, these priors warm-start under-explored arms while still letting the controller override poor priors based on observed GEO gains, as visualized by the bar charts in Figure~\ref{fig:bandit_results}. In practice, we run the controller for at most 10--15 iterations per item, after which it typically converges to a stable, high-reward tool choice, avoiding the need to evaluate all $K$ tools exhaustively.
\end{enumerate}

\begin{algorithm}[t]
\caption{Agent2UCB: Agent-Assisted Dual Upper Confidence Bounds}
\label{alg:llm-2ucb}
\begin{algorithmic}[1]
\Require Tool-selection agent $\mathbb{G}$, Content-analysis agent $\mathbb{A}$, target content $x$, horizon $T$, arms $K$, bandit hyperparameter $\alpha>0$, auxiliary size $n^{\text{aux}}$.

\State \textbf{Offline prediction:} Estimate $\bar{\mu}_k^{\text{agt}} = \Pr(\mathbb{G}(x,\mathbb{A}(x)) = k)$ for each $k\in[K]$, using $n^{\text{aux}}$ agent samples.

\State \textbf{Initialization:} For all $k\in[K]$, set $\bar{\mu}_k^{1}\gets 0$, $n_k^{1}\gets 1$, $s_k^{1}\gets 0$.

\For{$t=1$ to $T$}
  \State \textbf{First UCB:} $U_k^{1} = \bar{\mu}_k^{t} + \alpha\sqrt{\tfrac{\log t}{n_k^{t}}}$.
  \State \textbf{Second UCB:} $U_k^{2} = \dfrac{s_k^{t} + \bar{\mu}_k^{\text{agt}}\, n_k^{\text{aux}}}{n_k^{t}+n_k^{\text{aux}}}
      + \alpha\sqrt{\tfrac{\log t}{\,n_k^{t}+n_k^{\text{aux}}}}$.
  \State \textbf{Action:} $a_t = \arg\max_{k\in[K]} \min\{U_k^{1}, U_k^{2}\}$.
  \State \textbf{Observe} reward $r_{a_t}$ and update:
  $n_{a_t}^{t+1} \gets n_{a_t}^{t}+1$, \;
  $s_{a_t}^{t+1} \gets s_{a_t}^{t}+r_{a_t}$, \;
  $\bar{\mu}_{a_t}^{t+1} \gets \tfrac{s_{a_t}^{t+1}}{n_{a_t}^{t+1}}$.
\EndFor
\end{algorithmic}
\end{algorithm}

Together, these components implement a closed loop: baseline GEO and EEAT-R analysis diagnose the current state of a page, the bandit controller selects promising GEO tools from the library, the agentic engine generates optimized drafts, and the evaluation modules update both the reward signals and the user-facing dashboard. 

\section{Experiment}

We evaluate Agent2UCB against an unedited baseline and a category-level GEO baseline along three research questions: (RQ1) does the bandit controller reach higher generative visibility than the baselines, (RQ2) how many iterations does it need to get there, and (RQ3) do these visibility gains come at the cost of SEO quality.

Figure~\ref{fig:bandit_results} shows that Agent2UCB matches or exceeds both the category-level and exhaustive-search baselines on average GEO score. The right panel (RQ2) tracks GEO score against the number of bandit iterations: the dual-UCB controller converges within about 15 iterations, roughly $10\times$ fewer LLM calls than exhaustive search, while reaching comparable or higher scores. The bottom panel shows that no single tool dominates the tool-selection distribution across content items, confirming that the most effective optimization strategy varies by item rather than following one category-level rule -- the motivation for a per-item bandit controller in the first place.

\begin{table}[t]
\centering
\caption{Comparison of SEO quality before and after optimization under the Category baseline and Agent2UCB system.}
\begin{tabular}{lccc}
\toprule
\textbf{Metric} & \textbf{Original} & \textbf{Category} & \textbf{Agent2UCB} \\
\midrule
Overall SEO Score & 65.55 & 70.01 & 70.94  \\
Readability       & 0.810 & 0.762 & 0.718\\
Topical Coverage  & 0.551 & 0.800 & 0.653 \\
EEAT Credibility  & 0.426 & 0.396 & 0.773 \\
\midrule
Avg.\ Word Count  & 762.0 & 192.1 & 813.0  \\
Avg.\ Sentences   & 42.3  & 11.7  & 57.9  \\
\bottomrule
\end{tabular}
\label{tab:rq3_seo_metrics}
\end{table}

Table~\ref{tab:rq3_seo_metrics} addresses RQ3. Agent2UCB attains the highest Overall SEO Score of the three settings and raises EEAT credibility by roughly 80\% over the unedited page and 95\% over the category baseline, while avoiding the aggressive truncation the category baseline exhibits (192.1 vs.\ 813.0 average words). The trade-off is a modest readability dip relative to both the original page and the category baseline, and a topical-coverage score that improves over the original but falls short of the category baseline's.

\begin{figure}[t]
    \centering

    % ----- Top image -----
    \begin{subfigure}{\columnwidth}
        \centering
        \includegraphics[width=\linewidth]{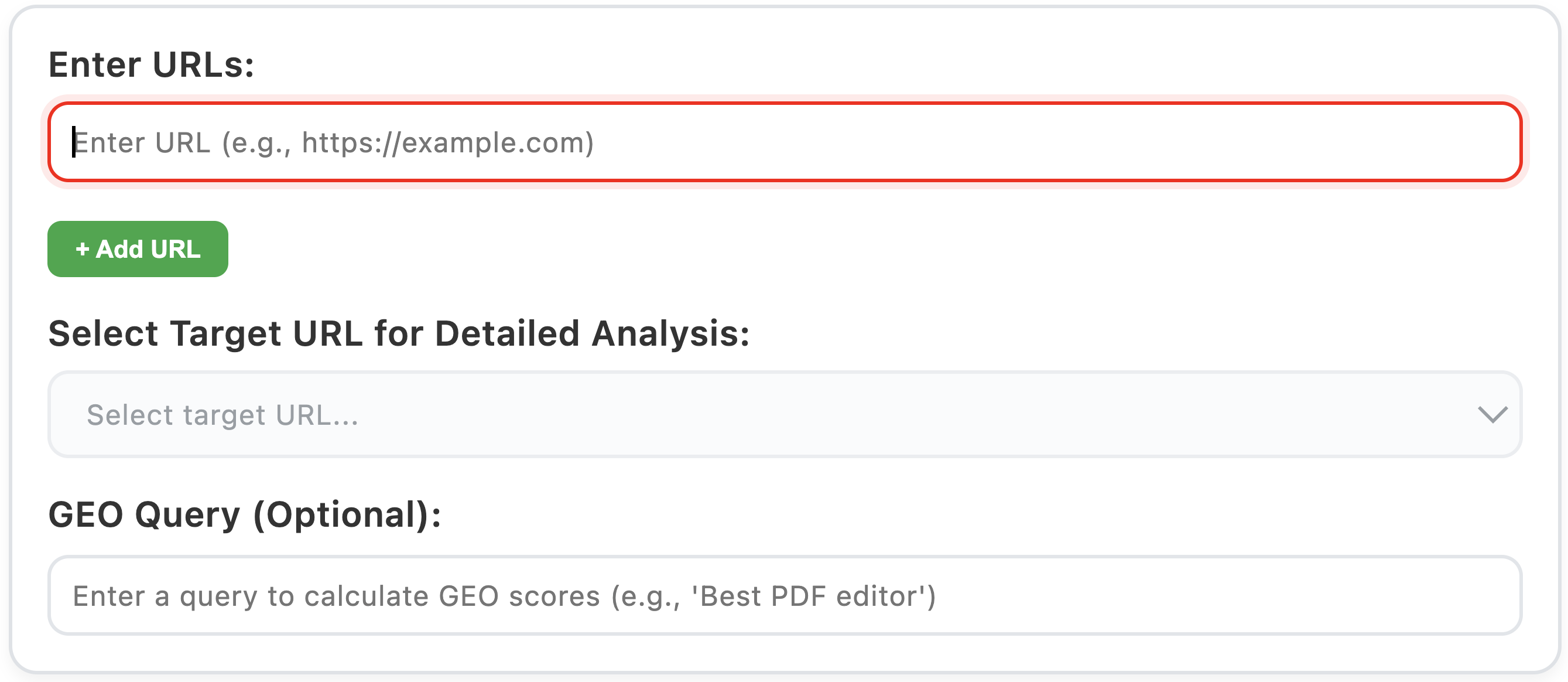}
    \end{subfigure}

    \vspace{0.6em} % spacing between images

    % ----- Bottom image -----
    \begin{subfigure}{\columnwidth}
        \centering
        \includegraphics[width=\linewidth]{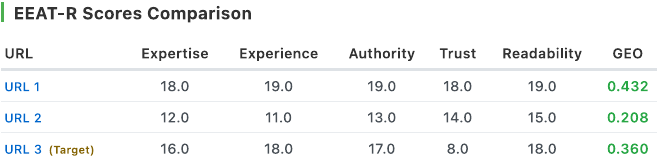}
    \end{subfigure}

    \caption{Agent2UCB dashboard: query/URL input view (top) and results comparison table (bottom).}
    \label{fig:combined2}
\end{figure}

\begin{figure}[t]
    \centering

    % ----- Top image -----
    \begin{subfigure}{\columnwidth}
        \centering
        \includegraphics[width=\linewidth]{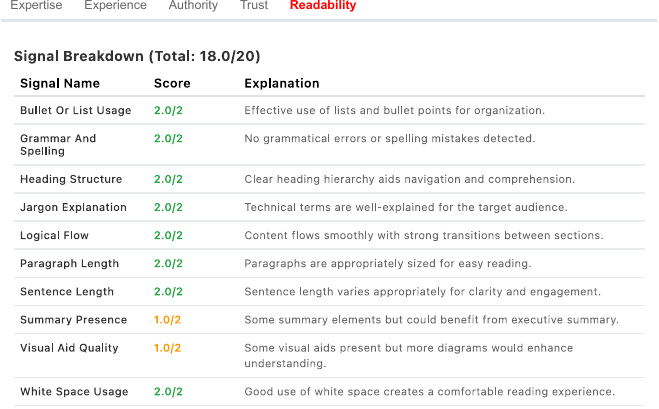}
    \end{subfigure}

    \vspace{0.6em} % spacing between images

    % ----- Bottom image -----
    \begin{subfigure}{\columnwidth}
        \centering
        \includegraphics[width=\linewidth]{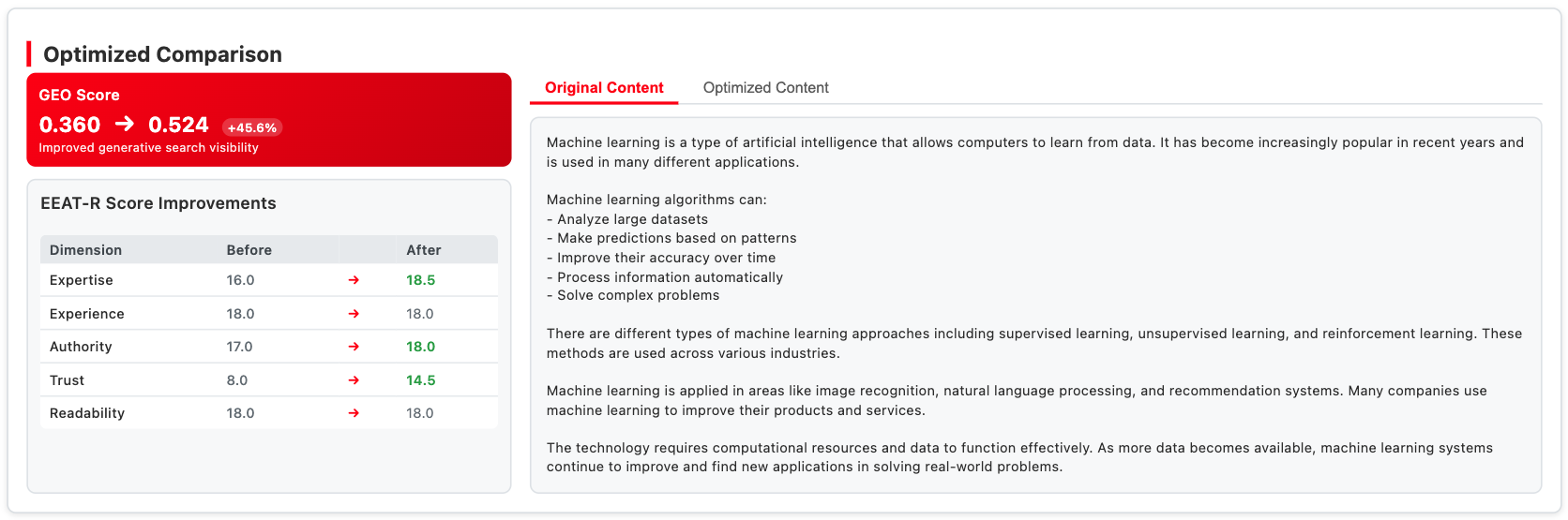}
    \end{subfigure}

    \caption{Agent2UCB dashboard: EEAT-R analysis view (top) and optimization result view (bottom).}
    \label{fig:combined}
\end{figure}

Figures~\ref{fig:combined2} and~\ref{fig:combined} show the demo interface end to end: users submit a query and candidate URLs and review a summary table of GEO/SEO outcomes across methods (Figure~\ref{fig:combined2}), then drill into the EEAT-R analysis for a chosen page and compare its optimized draft (Figure~\ref{fig:combined}).

\section{Conclusion}

This work introduces a bandit-based framework for \textbf{Generative Engine Optimization (GEO)},
formulating tool selection as an adaptive exploration–exploitation process rather than an
exhaustive search. By integrating \emph{multi-armed bandits} with verifiable visibility rewards,
our method significantly reduces computational cost while maintaining competitive performance.
The proposed \emph{Agent2UCB} controller further mitigates cold-start issues by combining
offline predictions with online feedback. Experimental results demonstrate that
Agent2UCB converges to the optimal optimization method within a few iterations, enabling
scalable and efficient deployment in real-world generative search systems.

\bibliographystyle{IEEEtran}
\bibliography{main}

\end{document}